\documentclass{article}
\usepackage{tabu}

\usepackage{multirow}

\usepackage[english]{babel}
\usepackage[utf8x]{inputenc}
\usepackage{graphicx}
\usepackage{wrapfig}
\usepackage{lipsum}
\usepackage{float}

\usepackage{amsmath,amssymb}
\DeclareMathOperator*{\argmax}{arg\,max}

\renewcommand{\vec}[1]{\mathbf{#1}}

\usepackage[ruled,vlined]{algorithm2e}
\usepackage{graphicx}
\usepackage{subcaption}

\usepackage[final]{corl_2020} 

\title{Explicitly Encouraging Low Fractional Dimensional Trajectories Via Reinforcement Learning}

\author{
  Sean Gillen\\
  Department of Electrical Engineering\\
  University of California Santa Barbara
  United States\\
  \texttt{sgillen@ucsb.edu} \\
  \And
  Katie Byl\\
  Department of Electrical Engineering\\
  University of California Santa Barbara
  United States\\
  \texttt{katiebyl@ucsb.edu} \\
}

\begin{document}
\maketitle


\begin{abstract}

  A key limitation in using various modern methods of machine learning in developing feedback control policies is the lack of appropriate methodologies to analyze their long-term dynamics, in terms of making any sort of guarantees (even statistically) about robustness.  The central reasons for this are largely due to the so-called curse of dimensionality, combined with the black-box nature of the resulting control policies themselves. This paper aims at the first of these issues. Although the full state space of a system may be quite large in dimensionality, it is a common feature of most model-based control methods that the resulting closed-loop systems demonstrate dominant dynamics that are rapidly driven to some lower-dimensional sub-space within. In this work we argue that the dimensionality of this subspace is captured by tools from fractal geometry, namely various notions of a fractional dimension. We then show that the dimensionality of trajectories induced by model free reinforcement learning agents can be influenced adding a post processing function to the agents reward signal. We verify that the dimensionality reduction is robust to noise being added to the system and show that that the modified agents are more actually more robust to noise and push disturbances in general for the systems we examined.

\end{abstract}



\section{Introduction}
   
    The availability of computation as a resource has been growing exponentially since at least the 1970s, and there is every indication that this resource will continue to become cheaper and more available well into the conceivable future. Researchers have been able to leverage the large amounts compute available to better control robotic systems, and advances in computational capacity and algorithmic development continue to open up new domains. One promising manifestation of this is model-free reinforcement learning, a branch of machine learning which allows an agent to interact with its environment and autonomously learn how to maximize some measure of reward. The promise here is to allow researchers to solve problems for systems that are hard to model, and/or that the user doesn't know how to solve themselves.  Recent examples in the context of robotics include controlling a 47 DOF humanoid to navigate a variety of obstacles \cite{heess_emergence_2017}, dexterously manipulating objects with a 24 DOF robotic hand \cite{openai_learning_2018}, and allowing a physical quadruped robot to run \cite{hwangbo_learning_2019}, and recover from falls \cite{lee_robust_2019}.
    
    In this paper we study legged locomotion. This class of problems is notoriously difficult, and as a result reinforcement learning is a popular tool to throw at it. We would argue that hand-designed, model based control still represents the state of the art (a la Boston Dynamics), but RL has been a fruitful approach. There are examples of learned policies outperforming hand designed ones \cite{hwangbo_learning_2019}, and there is good reason to believe these learning methods will continue their current trajectory of increasing performance gains and ease of use. But these algorithms have a serious draw back in that they are mostly black boxes. It is an open challenge to figure out what exactly it is that your RL agent has learned. If all you know is that one of your agents achieved very high reward, it is not clear how to verify that this system is safe and sensible in all the regions of state space it will visit during its life. Nor can we necessarily say anything about the stability or robustness properties of the system. Recent work ~\citep{Taleledeep} has used so-called mesh-based tools to examine precisely these questions. 
    
    However, utility of any mesh-based tool to accurately discretize a state-space is limited, due to the curse of dimensionality. In practice, these methods are only able to work on relatively high dimensional systems if the reachable space grows at a rate that is much smaller than the exponential growth of the full state space the system within which it is embedded. To expand these methods to higher dimensional systems. We will need to find ways to keep the volume of visited states from expanding commensurately. One way to quantify this rate of growth is by using one of the several notions of "fractional dimensions" from fractal geometry.
    
    In this work, we discuss an efficient meshing algorithm, which we call box meshing. We show that this approach makes calculating the so called mesh dimension feasible in the context of reinforcement learning. We also propose using other notions of fractional dimension from the literature as a proxy for the property we care about. We then show that reinforcement learning agents can be trained to shrink these measures by post processing their reward function. We present the results of this training, and finally present some brief analysis of the resulting structure for select policies.

\section{Meshing \& Fractional Dimensions}

\begin{figure}[h!]
  \centering
  \begin{subfigure}[b]{0.32\linewidth}
    \includegraphics[width=\linewidth]{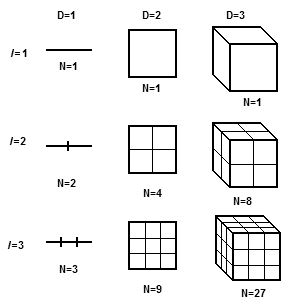}
    \caption{Scaling in different dimensions}
  \end{subfigure}
  \begin{subfigure}[b]{0.32\linewidth}
    \includegraphics[width=\linewidth]{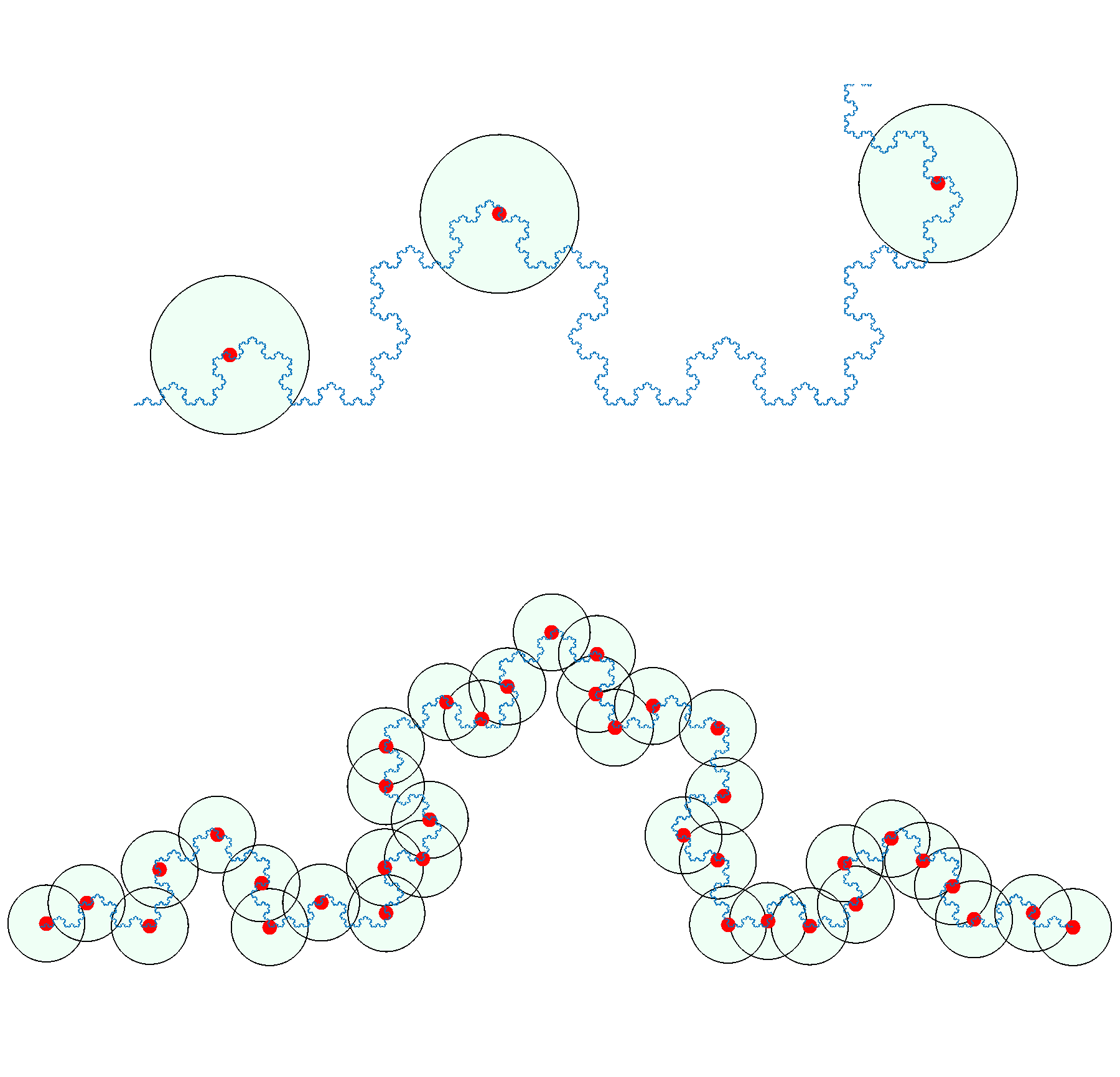}
    \caption{A non uniform mesh}
  \end{subfigure}
  \begin{subfigure}[b]{0.32\linewidth}
    \includegraphics[width=\linewidth]{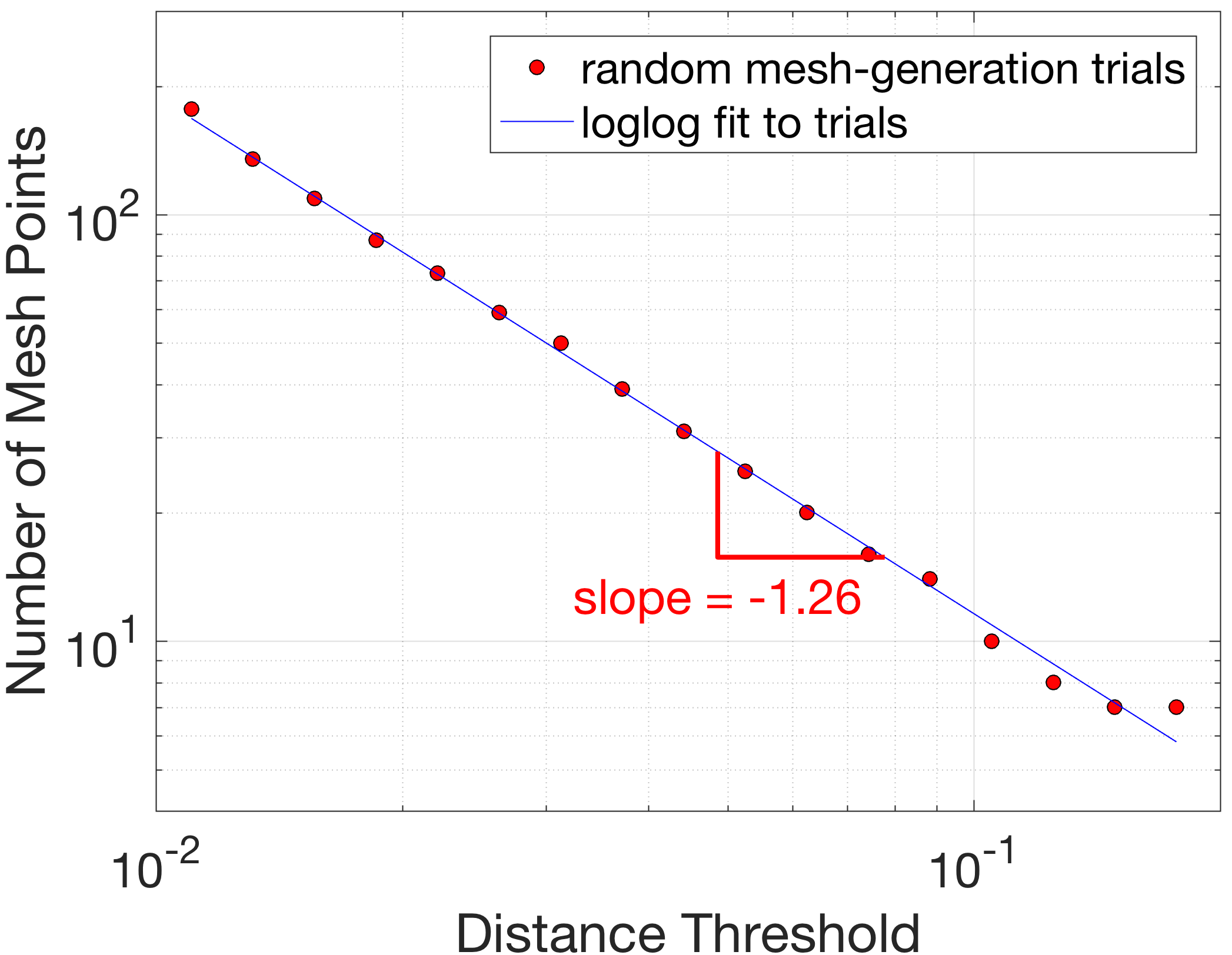}
    \caption{Calculating the mesh dimension}
  \end{subfigure}
  \caption{Image credit for sub-figure a: \cite{BrendanRyan/Publicdomain2020}, image credit for sub-figures b and c: \cite{Talelepush}}
  \label{fig:fracdim}
\end{figure}


Let's say we have a continuous set S that we want to approximate by selecting a discrete set M  composed of regions in S. We will call this set M a mesh of our space. Figure \ref{fig:fracdim}(a) shows some examples of this: a line is broken into segments, a square into grid spaces, and so on. The question is: as we increase the resolution of these regions, how many more regions N do we need? Again, figure \ref{fig:fracdim}(a) shows us some very simple examples. For a D dimensional system, if we go from regions of size d to d/k, then we would expect the number of mesh points to scale as $N \propto k^{D}$. But not all systems will scale like this, as \ref{fig:fracdim}(b) and \ref{fig:fracdim}(c) illustrate. Figure \ref{fig:fracdim}(b) is an example of a curve embedded in a two dimensional space. The question of how many mesh points are required must be answered empirically. Going backwards, we can use this relationship to assign a notion of "dimension" to the curve. 

\begin{equation}
    D_{f} = -\lim_{k \rightarrow 0}\frac{\log N(k)}{\log k} 
\end{equation}

What we are talking about is called the Minkowski–Bouligand dimension, also known as the box counting dimension. This dimension need not be an integer, hence the name "fractional dimension". As a practical matter, we use the slope of the log-log plot of mesh sizes over d to calculate this, rather than taking a limit. This is one of many measures of "fractional dimension" that that emerged from the study of fractal geometry. Although these measures were invented to study fractals, they can still be usefully applied to non-fractal sets.

In \cite{OguzSaglam2015}, Saglam and Byl introduced a technique that is able to simultaneously build a non-uniform mesh of a reachable state space while developing robust policies for a bipedal walker on rough terrain. Having a discrete mesh allows for value iteration over several candidate controllers, which found a robust control policy. In addition this mesh allows for the construction of a state transition matrix, which was used to calculate the mean first passage time \cite{Byl2009}, a metric that quantifies the expected number of steps a meta-stable system can take before falling. 

Since its introduction, meshing in this fashion has been used for designing walking controllers robust to push disturbances \cite{Talelepush}, to design agile gaits for a quadruped \cite{Byl2017}, and to analyze hybrid zero dynamics (HZD) controllers \cite{Saglam2015}. There has also been recent work to use these tools to analyze policies trained by deep reinforcement learning \cite{Taleledeep}. A long term goal and motivation for this work is to take a high performance controller obtained via reinforcement learning, and extract from it a mesh-based policy that is both explainable and amenable to analysis.

\subsection{Box Meshing}
\label{sec:boxmsh}
Our primary improvement to the prior work on meshing is to introduce something we call box meshing. Prior, a new mesh point could take any value in the state space. To determine if a new state is already in the mesh, we would compute a distance metric to every point in the mesh, and check if the minimum was below our threshold. Thus, building the mesh was an $O(n)^{2}$ algorithm. By contrast, in box meshing we apriori divide the space uniformly into boxes with side length $d$. We identify any state s with a key obtained by: $\text{key} = \text{round}(\frac{s}{d})d$, where round performs an element-wise rounding to the nearest integer. We can then use these keys to store mesh points in a hash table. Using this data structure, we can still store the mesh compactly, only keeping the points we come across. However, insertion and search are now $O(1)$, and so building the mesh is $O(n)$. This is very similar to non-hierarchical bucket methods, which are well studied spatial data structure \cite{Samet1990}, although we are using them for data compression here. In the prior meshing work, this sort of speedup would be minor, the run-time is dominated by the simulator or robot. However, this speedup does open some new possibilities: most poignantly, it makes calculating the mesh dimension during reinforcement learning plausible.

\subsection{Algorithmic Box Mesh Dimension}
\label{sec:boxdim}
The "mesh dimension" is the quantity extracted from the slope of the log log plot of mesh sizes vs d values. For this paper, it is assumed that the mesh algorithm being used for this calculation is the box mesh. Automatically computing the mesh dimension of a data set generated from learning agent with speed, accuracy, and robustness is very challenging. A single trajectory provides only a small amount of data, which adds significant noise to the mesh sizes. Agents might do things like fall over and generate extremely short trajectories, or learn a trajectory that "stands in place", which can lead to numerical errors. Finally, every decision is a trade-off between accuracy and speed. Model-free RL is predicated on having a huge number of rollouts to learn from, and we would like for any mesh-dimension quantification algorithm to be fast enough so as to not dominate the total learning time. With these factors in mind, we introduce two box mesh dimensions. The lower mesh dimension does the linear fit, but intentionally errs on the side of including flat parts of the graph, and therefore tends to underestimate the true mesh dimension. We then have the upper mesh dimension, which takes the largest slope in the log log relationship, thus tending to overestimate the true mesh dimension. Neither of these measures are correct, but taken together they can bound the mesh dimension, and as we will see they can be useful on their own.





\begin{algorithm}
\SetAlgoLined
\textbf{Input:} State set $S$, box size d. \\
\textbf{Output:} Mesh size m.\\
\textbf{Initialize:} Empty hash table M. \\
 \For{s $\in$ S}{
    $\bar s  = \text{Normalize(s)} $ \\
    key = round($\bar s$ / d)d \\
    \eIf{s $\in$ M}{
        M[key]++
    }{
        M[key]=1
    }
}
\textbf{Return:} M
\caption{Create Box Mesh, see section \ref{sec:boxmsh}}
\end{algorithm}

\begin{algorithm}
\SetAlgoLined
\textbf{Input:} State set $S$.\\
\textbf{Output:} Mesh M. \\
\textbf{Hyperparameters:} scaling factor f,  initial box size $d_{0}$. \\
\textbf{Initialize:} Empty list of mesh sizes H, empty list of d values D. \\
m = Size(CreateBoxMesh(S, $d_{0}$)) \\
d = $d_{0}$ \\ 
Append m to H, append d to D. \\
\While{m $<$ size(S)}{
    d = d/f \\ 
    m = Size(CreateBoxMesh(S, d)) \\ 
    Prepend m to H, prepend d to D. \\
}
\While{m $\neq$ 1}{
    d = d*f \\ 
    m = Size(CreateBoxMesh(S,d)) \\
    Append m to H, append d to D. \\
}

X =  $\log$ d\\
Y =  -$\log$ m

\textbf{Lower Mesh Dim:} fit Y = gX + b, \textbf{Return:} g \\
\textbf{Upper Mesh Dim:} w = greatest slope in Y over X \textbf{Return:} w

\caption{Compute Box Mesh Dimension, see section \ref{sec:boxdim}}
\label{algo:mesh_dim}
\end{algorithm}


\section{Reinforcement Learning}
The goal of reinforcement learning is to train an agent acting in an environment to maximize some reward function. At every timestep $t \in \mathbb{Z}$, the agent receives the current state $s_{t} \in R^{n}$, uses that to compute an action $a_{t} \in \mathbb{R}^{b}$, and the receives the next state $s_{t+1}$, which is used to calculate a reward $r : \mathbb{R}^{n} \times \mathbb{R}^{m} \times \mathbb{R}^{n} \rightarrow \mathbb{R}$. The objective is to find a policy  $\pi_{\theta}: \mathbb{R}^{n} \rightarrow \mathbb{R}^{m}$

\begin{equation} \argmax_{\theta} \mathop{\mathbb{E}}_{\eta}\left[ \sum_{t=0}^{T}r(s_{t}, a_{t}, s_{t+1}) \right] \end{equation}

Where $\theta \in \mathbb{R}^{d}$ is a set that parameterizes the policy, and $\eta$ is a parameter representing the randomness in the environment. This includes the random initial conditions for episodes.

\subsection{Post Processing Rewards}

In order to influence the dimensionality of the resulting policies, we introduce various postprocessors, which act on the reward signals before passing them to the agent. These obviously modify the problem: in some sense the postprocessed environment is a completely different problem from the original. However our meta-goal is to train agents that achieve reasonable rewards in the base environment, while simultaneously exhibiting reduced dimensionality we are looking for. These postprocessors take the form:
\begin{equation}
R_{*}(\vec{s}, \vec{a}) = \frac{1}{D_{*}(\vec{s})}\sum_{t=0}^{T} r(s_{t}, a_{t}, s_{t+1})
\label{post}
\end{equation}

Where $\vec{s}, \vec{a}$ are understood to be an entire trajectory of state action pairs, and $D_{*}$ is some measure of fractional dimension. Some measures of dimensionality can be inserted here directly (See \ref{sec:var}). However the mesh dimensions computed by algorithm \ref{algo:mesh_dim} require a little more care. We must first define a clipped dimension:

\begin{equation}
D^{c}_{*} = \text{clip}[D_{*}(\vec{s}_{t > Tr}), 1, D_{t}/2)] 
\end{equation}

where $D_{t}$ is the topological dimension, equal to the number of states in the system. $Tr$ is a fixed timestep chosen to exclude the initial transients resulting from a system moving from rest to into a quasi-cyclical ``gait''. In this paper we set $Tr$ = 200 for all experiments. For comparison, the nominal episode length is 1000. The clipping is to ensure that the pathological trajectories that and RL agent sometimes generates don't interfere with the training. It will also clip trajectories that terminate early, to prevent agents learning to fall over immediately to ``game the system''. Half of the topological dimension proved to be a decent upper bound for the worst case dimensionality of each system in practive. The \textbf{mesh dimension postprocessors} use the clipped dimension. Finally, when $D_{*} = 1$ is used, we call the result is the \textbf{identity post processor}, since in this case the total reward is completely unchanged.

\subsection{Environments}
We examine a subset of the popular OpenAI Mujoco locomotion environments introduced in \cite{1606.01540}. In particular, we evaluate our work on HalfCheetah-v2, Hopper-v2, and Walker2d-v2. These environments were chosen because they have a relatively high dimensionality (11-17 DOF), yet we believe can be made feasible for meshing based approaches. The state space consists of all joint / base positions and velocities, with the x (the "forward") position being held out, because we want a policy that is invariant along that dimension.


\subsection{Augmented Random Search}

 In \cite{Mania2018} Mania et al introduce Augmented Random Search (ARS) which proved to be efficient and effective on the locomotion tasks. Rather than a neural network, ARS used static linear policies, and compared to most modern reinforcement learning, the algorithm is very straightforward. The algorithm operates directly on the policy weights, each epoch the agent perturbs it's current policy N times, and collects 2N rollouts using the modified policies. The rewards from these rollouts are used to update the current policy weights, repeat until completion. The algorithm is known to have high variance; not all seeds obtain high rewards, but to our knowledge their work in many ways represents the state of the art on these benchmarks. Mania et al introduce several small modifications of the algorithm in their paper, our implementation corresponds to the version they call ARS-V2t.






\subsection{Training}
\label{sec:training}

To keep things simple, we wanted to find one set of hyper parameters for all environments and post processors. These parameters were chosen by hand, with the parameters reported in Table 9 from \cite{Mania2018} as a starting point. We tuned until our unprocessed learning achieved satisfactory results across all tasks. Again, ARS is known to have high variance between random seeds, and some seeds never learn to gather a large reward. The parameters we found are able to consistently solve the cheetah and walker; for the hopper, the algorithm learns a policy with high reward around half the time. This seems consistent with the performance reported in \cite{Mania2018}. We train each postprocessor on 10 random seeds, the evaluation metrics are averages over 5 rollouts from each seed, and for the dimension metrics we use extended episodes of length 10,000 to get a more accurate measurement. The reported returns, and the training, both use the normal 1,000 step episodes. We found that the mesh postprocessors were getting very poor performance when trained from a random policy. However, we found that we saw good results when these trials were initialized with a working policy. Therefore we trained agents for 750 epochs without post processing, and used that to initialize the mesh dimension policies. The mesh policies were then trained for an additional 250 epochs and the results reported.

\section{Results}
\subsection{Mesh Dimension Postprocessors}

\begin{figure}[!htb]
  \centering
  \begin{subfigure}[b]{0.32\linewidth}
    \includegraphics[width=\linewidth]{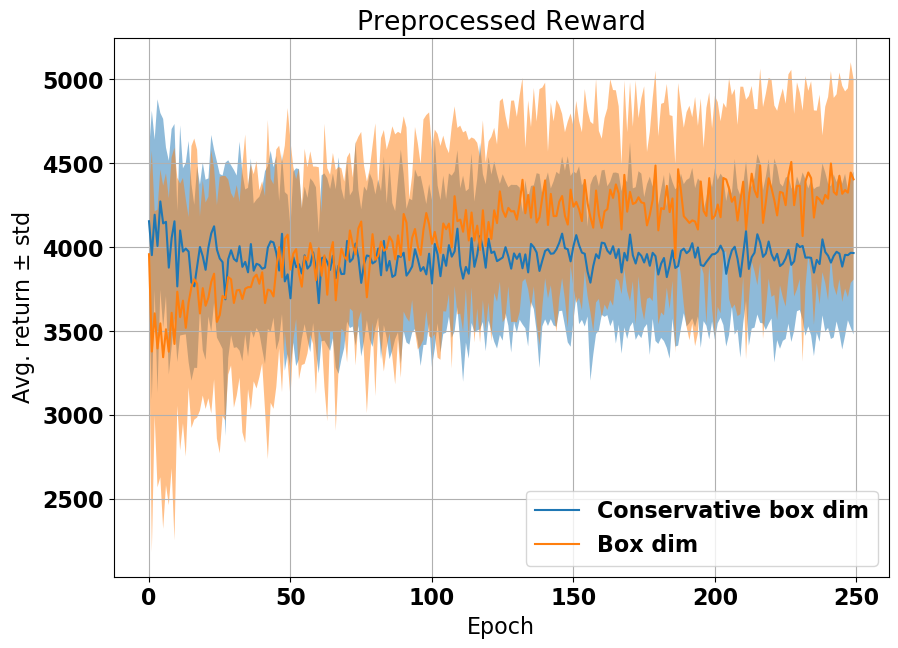}
    \caption{HalfCheetah}
  \end{subfigure}
  \begin{subfigure}[b]{0.32\linewidth}
    \includegraphics[width=\linewidth]{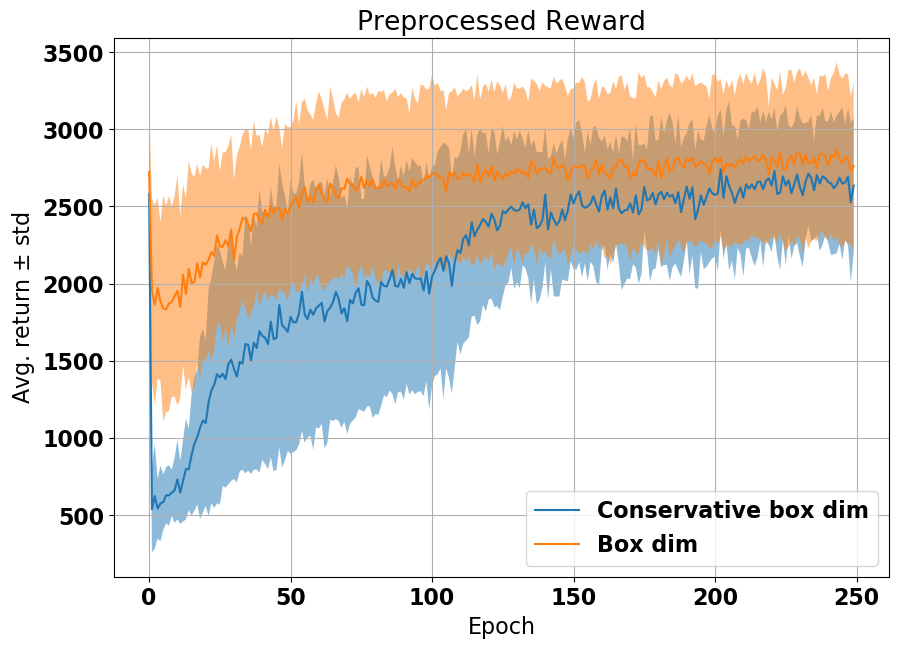}
    \caption{Hopper}
  \end{subfigure}
  \begin{subfigure}[b]{0.32\linewidth}
    \includegraphics[width=\linewidth]{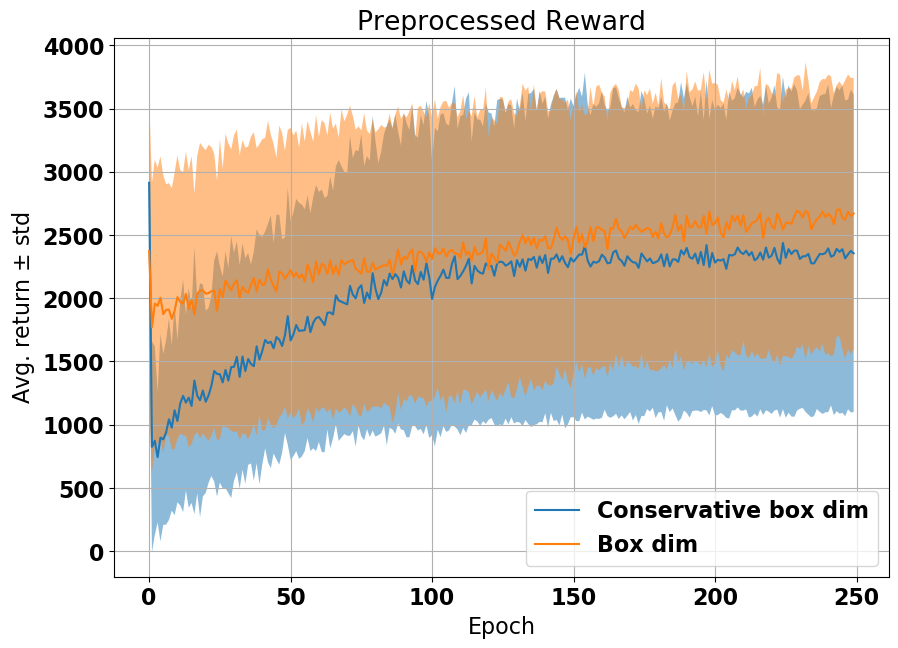}
    \caption{Walker}
  \end{subfigure}
  \caption{Reward curves for mesh dimension postproccesor runs.}
  \label{fig:mesh_rews}
\end{figure}

\begin{table}[!htb]
\begin{tabular}{ l|l|l|l|l }
\hline
Environment & Postprocessor 
                  & Lower Mesh Dim.         & Upper Mesh Dim.   & Return \\ 
\hline
\multirow{3}{2.6cm}{HalfCheetah-v2} 
& Identity        & 2.31 $\pm$ 0.71   & 7.34 $\pm$  1.56 &  5469 $\pm$ 823 \\
& Lower Mesh Dim        &  \textbf{0.66 $\pm$ 0.51}   & \textbf{2.55 $\pm$  1.52} &  4962 $\pm$ 598 \\
& Upper Mesh Dim.  & \textbf{1.06 $\pm$ 1.13}  & \textbf{2.83  $\pm$ 1.27} &  4432 $\pm$ 539 \\
\hline
\multirow{3}{2.6cm}{Hopper-v2} 
& Madogram$^{*}$        & 1.62 $\pm$ .27   & 4.68  $\pm$  0.82   &  3461 $\pm$ 119 \\
& Lower Mesh Dim.        & 1.13 $\pm$ .02   & 3.54 $\pm$  0.96   &  2941 $\pm$ 538 \\
& Upper Mesh Dim.  & 1.27 $\pm$ .50   & 2.98 $\pm$  1.48   &  3020 $\pm$ 337 \\
\hline
\multirow{3}{2.6cm}{Walker2d-v2 \\ (walking seeds)$^{**}$}
& Identity        & 2.13 $\pm$  0.31    & 4.62 $\pm$ 1.03  &  3758 $\pm$ 1037  \\
& Lower Mesh Dim.       & 1.21 $\pm$  0.06    & 4.09 $\pm$ 1.03  &  3339 $\pm$ 887   \\
& Upper Mesh Dim. & 1.89 $\pm$  0.42    & 3.10 $\pm$ 0.93  &  3359 $\pm$ 903 \\
\hline
\multirow{3}{2.6cm}{Walker2d-v2 \\ (all seeds)$^{**}$ }
& Identity        & 2.13 $\pm$ 0.31   & 4.62 $\pm$ 1.03    &  3758 $\pm$ 1037 \\
& Lower Mesh Dim.       & 1.04 $\pm$ 0.53   & 4.45 $\pm$ 1.19    &  3034 $\pm$ 1086 \\
& Upper Mesh Dim. & 1.48 $\pm$ 0.67   & 2.27 $\pm$ 0.95    &  2556 $\pm$ 1378 \\
\hline
\end{tabular}
\caption{\label{tab:mesh} Mesh dimensions and returns for trajectories after training. See \ref{sec:training} for details\\
\footnotesize{*  Because ARS with our chosen hyper parameters does not consistently produce 10 seeds that perform well on the hopper, we instead use madodiv (see the \ref{sec:var}) for the seed policies.  \\
**  See \ref{sec:msh}}
}
\end{table}

For all environments the mesh post processors had a significant impact on the mesh dimensions. It's important to remember here, the dimensions reported represent lower and upper bounds for the actual mesh dimensions. There was also a corresponding and significant decrease in the unprocessed rewards. However with our meta goal of training agents that have acceptable reward but which are more amenable to meshing, this is a more than acceptable trade. In the case of walker, several seeds (4 for the upper dim., 3 for the lower mesh dim.), "forget how to walk", and learn a policy that stands in place. This certainly has a low dimensionality, but is not very useful, to be complete we include statistics from the seeds that learned a gait, and for all 10 seeds, including the standing policies.

\label{sec:msh}

\section{Analysis}

We now examine the learned behavior for one of the more notable policies. By far the most dramatic effect from the tables above was the mesh dimension postprocessors on the cheetah. Both measures of dimension shrunk by 2-4 times.
Figure 7 presents data for this case.



\begin{figure}[!htb]
    \centering
    \includegraphics[width=\textwidth]{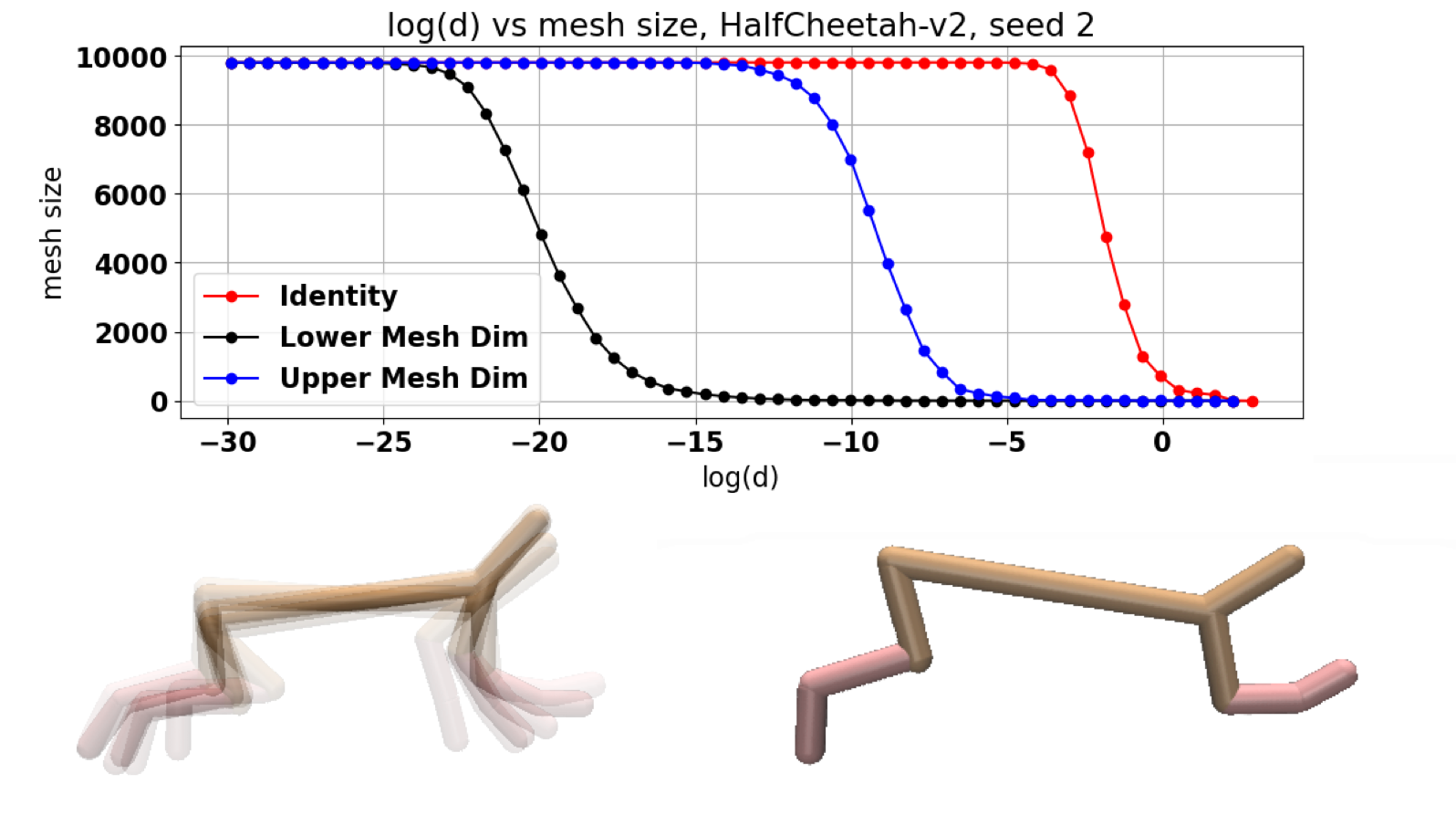}        
    \caption{Top: mesh sizes vs log of the box size for the cheetah environment. Lower Left: Every five frames overlaid for the an identity policy on the cheetah. Lower Right: Every five frames of cheetah after the lower mesh dimension training.}
    \label{fig:mesh_anal}
\end{figure}


Toward more intuitively understanding this data, a few comments are worth making, first. We have discussed the mesh dimension rather abstractly so far. In visualizing what this really means, imagine two different gait cycles. In one case, there is a general pattern to the motion, but it wanders in a noisy-looking way, like a ``signature'' that does not quite match up, cycle after cycle.  As motions become closer to being exact limit cycles, there is a more clear pattern of repetition, exactly analogous to re-tracing the same path, again and again, within the state space. Such a more tightly-structured limit cycle nature in turn results in a significantly lower-dimensional set of states being visited. 

We can see from the mesh size curves in Figure 7 that there is an overwhelming difference in the mesh sizes between the lower mesh dimension post processor and the other two. To put this in perspective, before the extra 250 epochs of training, if given a box size of .01, the agent would need a unique mesh point for every single state in the 10,000 state trajectory. After the additional training, the agent can represent all 10,000 points with just 5 mesh points! In this case it appears both agents learned a quasi periodic gait with period 5. In figure \ref{fig:mesh_anal} we present an overlay of the agents rendered every 5 steps. The results show us that the mesh agent has learned an extremely tight limit cycle. It's a bit of a strange limit cycle, being only 5 timesteps long, but nonetheless we think this is interesting and surprising behavior.







The behavior displayed in figure \ref{fig:mesh_anal} is clearly something that can only happen in a noiseless simulation, so we also measured the mesh dimensions of our policies when subjected to noise during rollouts using the noise values from the robustness experiments (see table \ref{tab:robust})to inform parameter values. The difference in fractal dimension is less pronounced than the no noise case but is still a clear improvement. Furthermore we expect that if we were to add noise at training time, that the learning may be able to find ways to lower the mesh dimension that is more robust to noise.

\begin{table}
\begin{tabular}{ l|l|l|l|l }
\hline
Environment & Postprocessor 
                  & Lower Mesh Dim.         & Upper Mesh Dim.   & Return \\ 
\hline
\multirow{3}{2.6cm}{HalfCheetah-v2} 
& Identity              &  2.38   $\pm$ 0.43   & 6.65 $\pm$  1.90 &  5404 $\pm$ 1015 \\
& Lower Mesh Dim        &  1.51   $\pm$ 0.13   & 3.03  $\pm$  1.09 &  4952 $\pm$ 572 \\
& Upper Mesh Dim.       & 1.76    $\pm$ 0.53      & 3.54  $\pm$  1.27 &  4222 $\pm$ 803 \\
\hline
\multirow{3}{2.6cm}{Hopper-v2} 
& Madogram$^{*}$         & 1.63 $\pm$ .14   & 4.49 $\pm$  0.75   &  3438 $\pm$ 185 \\
& Lower Mesh Dim.        & 1.67 $\pm$ .22   & 3.71 $\pm$  0.89   &  2943 $\pm$ 535 \\
& Upper Mesh Dim.        & 1.64 $\pm$ .16   & 3.01 $\pm$  1.36   &  3019 $\pm$ 337 \\
\hline
\multirow{3}{2.6cm}{Walker2d-v2 \\ (walking seeds)$^{**}$}
& Identity        & 2.13 $\pm$  0.31    & 4.62 $\pm$ 1.03  &  3758 $\pm$ 1037  \\
& Lower Mesh Dim. & 1.83 $\pm$  0.34    & 2.73 $\pm$ 0.75  &  3511 $\pm$ 872   \\
& Upper Mesh Dim. & 1.60 $\pm$  0.33    & 4.01 $\pm$ 1.18  &  3384 $\pm$ 903 \\
\hline
\multirow{3}{2.6cm}{Walker2d-v2 \\ (all seeds)$^{**}$ }
& Identity        & 2.10 $\pm$ 0.34   & 4.42 $\pm$ 1.00    &  3743 $\pm$ 1034 \\
& Lower Mesh Dim. & 1.68 $\pm$ 0.70   & 4.19 $\pm$ 1.25    &  3048 $\pm$ 1071 \\
& Upper Mesh Dim. & 1.48 $\pm$ 0.38   & 2.98 $\pm$ 0.86    &  2558 $\pm$ 1373 \\
\hline
\end{tabular}
\caption{\label{tab:mesh} Mesh dimensions and returns for trajectories subject to zero mean Guassian noise. Standard deviation of .001 and .01 was added to all actions and observations respectively. See \ref{sec:training} for details\\
\footnotesize{Because ARS with our chosen hyper parameters does not consistently produce 10 seeds that perform well on the hopper, we instead use madodiv (see the \ref{sec:var}) for the seed policies.  \\
**  See \ref{sec:msh}}
}
\end{table}

We also performed analysis on the robustness properties of the resulting polices. We tested the failure rate for agents in the presence of noise and disturbances, figure \ref{tab:robust} shows the results. It's worth noting at this point that in practice the lower mesh dimension seems to work better in practice than the upper one. We found that when computing the mesh dimension by hand (by hand fitting a line to a set of carefully obtained mesh size data) that the hand picked value was generally much closer to the lower mesh dimension, at least for the three systems we studied. Training with the lower mesh dimension also resulted in agents that were more robust and achieved higher reward compared to the upper dimension.  

We tested three different cases, adding zero mean Gaussian noise to the actions and observations, and adding a force disturbance to the center of mass of the agents during their rollouts. For the push disturbances we have two parameters, the rate of disturbances, and the mangitude of the force applied. At every step we sample uniformly from [0,1], if the result is less than the rate parameter, then a force is applied at that timestep. The force is applied at a random angle in the xz plane, with the fixed magnitude from the magnitude parameter. For each type of disturbance we did a grid search over the parameters and report the parameter for which the identity post processor failed in around 20 percent of cases. Failure is defined as an early termination of an episode.

\begin{table}
\centering
\begin{tabular}{ l|l|l|l }

\hline
            & Identity      & Lower mesh dim. & Action std  \\ 
\hline
Cheetah     &          0.24  &           0.05 & .05 \\
Hopper      &          0.19  &           0.10 & .05 \\ 
Walker      &          0.28  &           0.03 & .15 \\
\hline
\hline
            & Identity       & Lower mesh dim. & Observation std  \\ 
\hline
Cheetah     &          0.20  &           0.02 & .005 \\
Hopper      &          0.20  &           0.25 & .02 \\ 
Walker      &          0.18  &           0.10 & .03 \\
\hline
\hline
            & Identity       & Lower mesh dim. & Magnitude, Rate  \\ 
\hline
Cheetah     &          0.21  &           0.03 & 3, .2 \\
Hopper      &          0.17  &           0.10 & 1, .2 \\ 
Walker      &          0.20  &           0.00 & 1, .2 \\
\hline
\end{tabular}
\caption{\label{tab:robust} Failure rates for agents under various noise and push disturbances}
\end{table}

    
    




\section{Conclusion}

In this work, we introduced a technique to influence the fractional dimension of the closed-loop dynamics of a system through the use of novel, dimensionality-based modifications to  the cost functions for reinforcement learning policies. We demonstrate this technique on several benchmark tasks, and we briefly analyze a resulting policy to verify the outcome, demonstrating a much smaller mesh dimension without a large loss in reward or function.

\label{sec:conclusion}




\clearpage



\bibliography{biblio}  

\clearpage

\section{APPENDIX}

\subsection*{Hyper Parameters}

\textbf{ARS:} For all environments $\alpha = .02$, $\sigma = .025$, $N=50$, $b=20$.  \\
\textbf{MeshDim:} f = 1.5, $d_{0}$ = 1e-2

\subsection{Variation Estimators}
\label{sec:var}
As discussed, computing the mesh dimension automatically is fraught with peril, in many practical scenarios. But there are also many other, different metrics one might consider, to give various approximations to the fractional dimension we seek to estimate. Gneiting et al.~\cite{Gneiting2012} compare a number of these estimators, and submit that the variation estimator \cite{Emery2005} offers a very good trade off between speed and robustness. To obtain this estimator, first define the power variation of order p as:

\begin{equation}
P_{p}(X, l) = \frac{1}{(2n-l)}\sum_{i=l}^{n} | X_{i} - X_{i-l}|^{p}
\end{equation}

Then, we define the variation estimator of order p as:

\begin{equation}
Dv_{p}(X) = 2 - \frac{\log P_{p}(X,2) - \log P_{p}(X, 1)}{p\log 2}
\label{eq:var}
\end{equation}

The \textbf{madogram} estimator is the special case of \eqref{eq:var} where p = 1, and the \textbf{variogram} is where p = 2.

\subsection{Variational Postprocessors}

\begin{figure}[!htb]
  \centering
  \begin{subfigure}[b]{0.32\linewidth}
    \includegraphics[width=\linewidth]{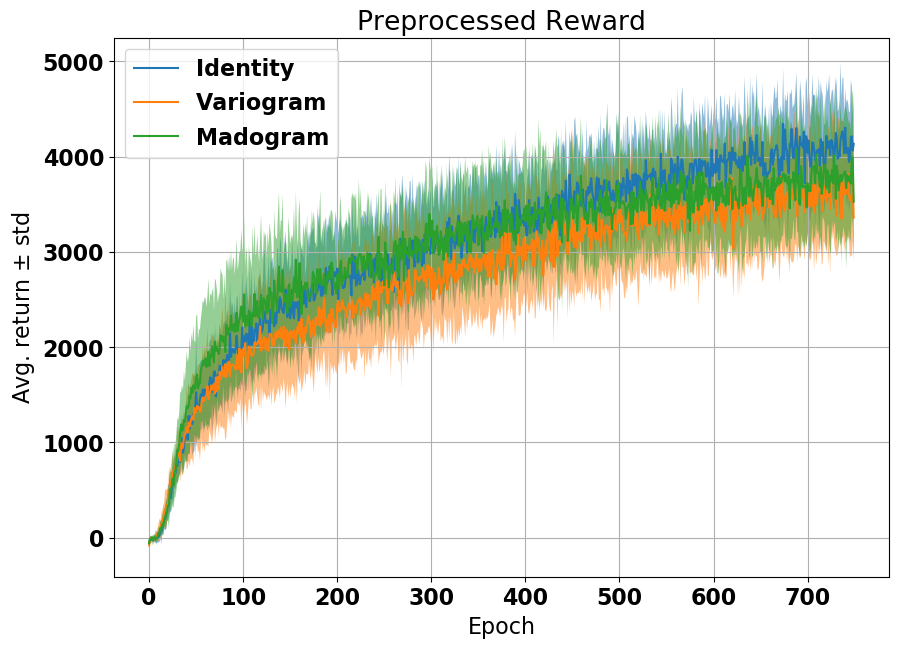}
    \caption{HalfCheetah}
  \end{subfigure}
  \begin{subfigure}[b]{0.32\linewidth}
    \includegraphics[width=\linewidth]{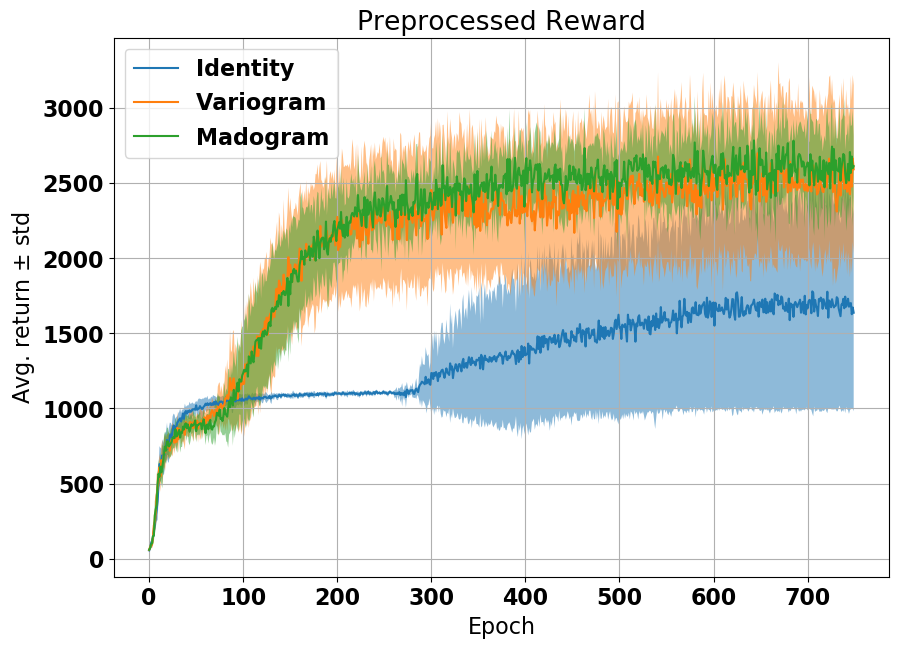}
    \caption{Hopper}
  \end{subfigure}
  \begin{subfigure}[b]{0.32\linewidth}
    \includegraphics[width=\linewidth]{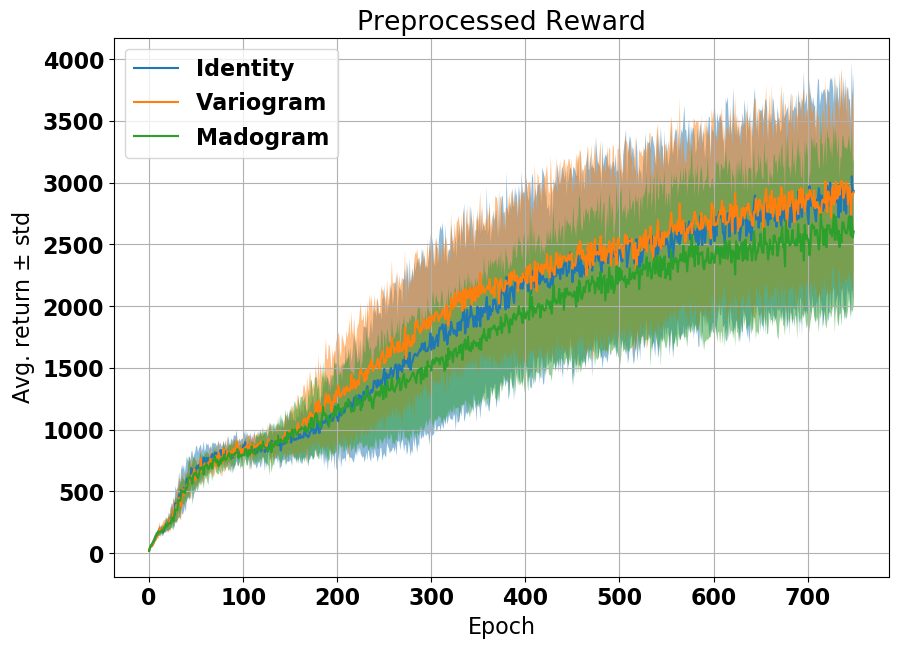}
    \caption{Walker}
  \end{subfigure}
  \caption{Reward curves for the variation postprocessors}
  \label{fig:rews}

\end{figure}
\begin{table}
\begin{tabular}{ l|l|l|l|l|l }
\hline
Environment & Postprocessor 
                    & Variogram        & Madogram       & Lower Mesh Dim.   & Return \\ \hline
\multirow{3}{*}{HalfCheetah-v2} 
& Identity          & 1.71 $\pm$ .03   & 1.42 $\pm$ .05 &  2.36 $\pm$ .61 & 5545 $\pm$ 593  \\
& Variogram   & 1.68 $\pm$ .01   & 1.36 $\pm$ .02 &  2.06 $\pm$ .60 & 5136 $\pm$ 851\\
& Madogram    & 1.65 $\pm$ .02   & 1.31 $\pm$ .04 &  2.09 $\pm$ .64 & 5234 $\pm$ 950\\

\hline
\multirow{3}{*}{Hopper-v2} 
& Identity$^{*}$           & 1.61 $\pm$ .14  & 1.22 $\pm$ .28               &  1.03$^{*}$ $\pm$ .71 & 2063 $\pm$ 1052 \\
& Variogram   &  \textbf{1.51 $\pm$ .02}  & \textbf{1.03 $\pm$ .04}   &  1.58 $\pm$ .54 & 3299 $\pm$ 711 \\
& Madogram     & \textbf{1.51 $\pm$ .002}  & \textbf{1.02 $\pm$ .004} &  1.57 $\pm$ .36 & 3449 $\pm$ 146\\
\hline
\multirow{3}{*}{Walker2d-v2} 
& Identity           & 1.68 $\pm$ .35  & 1.36 $\pm$ .71 &              2.14 $\pm$ .29 & 3742 $\pm$ 1038\\
& Variogram    & \textbf{1.54 $\pm$ .07}  & \textbf{1.07 $\pm$ .01} &  1.85 $\pm$ .54 & 3779 $\pm$ 894 \\
& Madogram     & \textbf{1.53 $\pm$ .01}  & \textbf{1.06 $\pm$ .02} &  1.99 $\pm$ .53 & 3414 $\pm$ 1025\\
\hline
\end{tabular}
\caption{\label{tab:dims} Mesh dimensions and returns for trajectories after training. See \ref{sec:training} for details\\
\footnotesize{* This includes policies which learned to "stand still", which lowers the average mesh dimension considerably see discussion}
}
\end{table}

It seems that the variational postprocessors had a modest effect the variational dimension, but that does not seem to correlate to a smaller mesh dimension, despite what our preliminary tests had led us to believe. The hopper and walker did have remarkable consistency in the variation dimensions they found; possibly this could be used to lower the variance in ARS. The fact that the variogram and madogram also got higher performance on the hopper task could support this claim. However without running many more trials and hyper parameter sweeps, that's not a claim that can be substantiated. These experiments show that 1) measures for fractional dimension can be influenced without adversely effecting the reward, and 2) that it is possible for an agent to shrink it's variogram and madogram dimensions without a large impact on its mesh dimension.

\subsection{Mesh Dimension Examples}

\begin{figure}
  \centering
  \begin{subfigure}[b]{0.45\linewidth}
    \includegraphics[width=\linewidth]{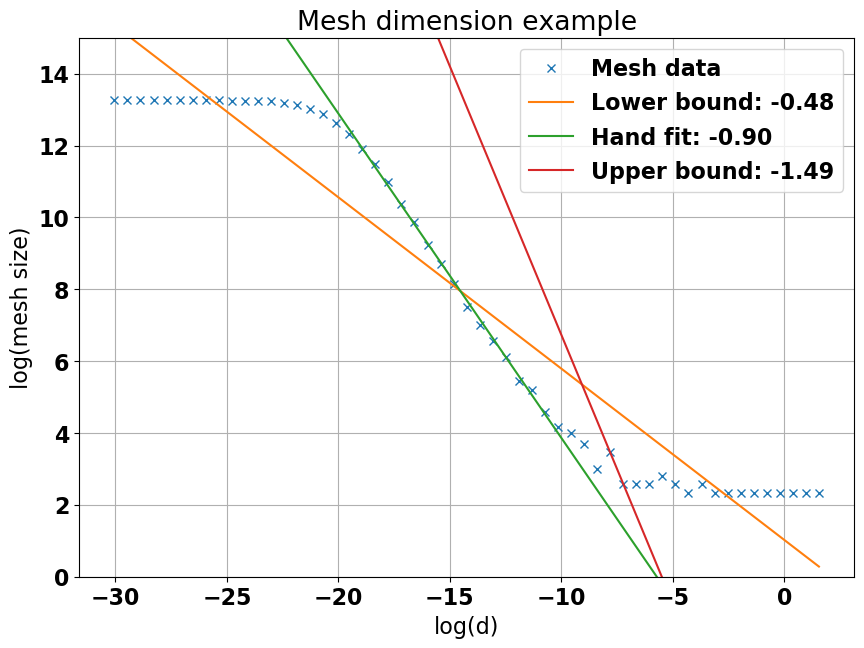}
    \caption{High resolution curve of a well behaved policy}
  \end{subfigure}
  \begin{subfigure}[b]{0.45\linewidth}
    \includegraphics[width=\linewidth]{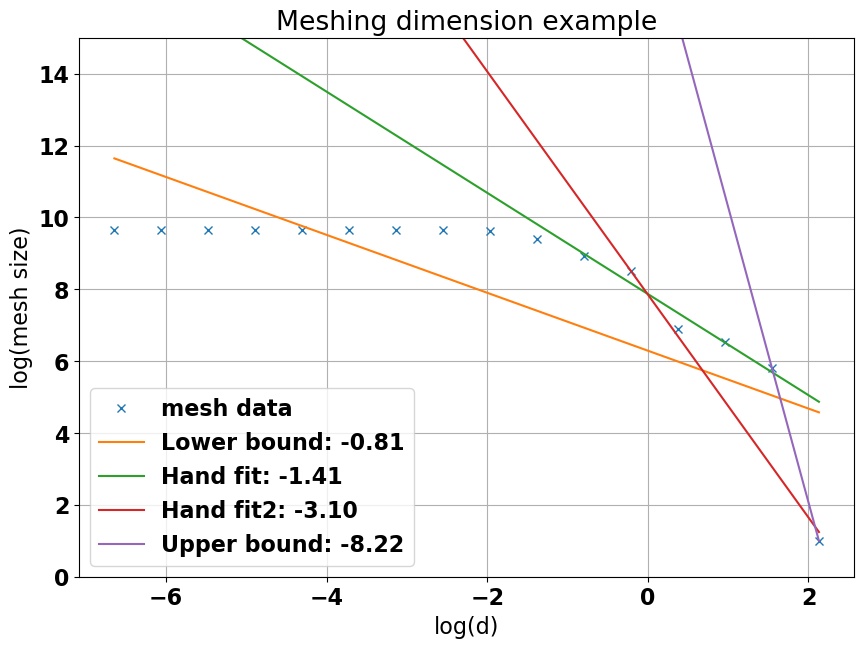}
    \caption{Run time resolution curve of a typical policy}
  \end{subfigure}
  \caption{Mesh curve and mesh dimension examples}
  \label{fig:curves}
\end{figure}

Figure \ref{fig:curves} illustrates two examples of the curves used to compute the mesh dimension. Recall that to compute the mesh dimension, we choose several values for d, the box length, and for each d construct a mesh using that box size. The x axis of these plots represents the log of the box length used, the y axis represents the log of size of the mesh created. For each curve, we display the lower bound and upper bound for the dimension as computed by algorithm 2, as well as several hand fits of the data. We hope that figure \ref{fig:curves}a makes clear what a close to ideal situation looks like, and provide intuition as to why the upper and lower mesh dimension bound the quantity we are trying to measure. Figure \ref{fig:curves}b serves to illustrate some of the problems with making an algorithmic measure of the dimension. There is much less data to work with due to performance constraints, which causes a large amount of noise on the estimate of the mesh dimension. Indeed even fitting this data by hand becomes a challenge and we provide two fits which can both be argued to be "correct". Which of these two represents the quantity we care about depends on the exact system being used and the purpose of the meshes we want to build with the resulting policy.

\subsection{Implementation Details}

For performance reasons, the mesh dimension algorithm does not actually create meshes until the mesh size equals the total data size, but rather until the mesh size is 4/5 the total data size. Figure 8a shows a typical mesh curve, and we can see the long tail of values with mesh sizes close to the maximum value. Not much useful information is gained from this and it is wasting time, so we stop early. We do not place the same limitation on the lower size of the mesh, since typically the mesh size hits one much more rapidly, again figure 8a illustrates this. In addition implement a minimum size for d, set to 1e-9 in this work to avoid numerical errors. 

The normalization done during box creation uses a running mean and standard deviation of all states seen so far during training. These stats are saved and used for evaluation as well, we found that the upper mesh dimension is very sensitive to the normalization used, but that the other metrics where not.

\end{document}